\relax
\documentclass[letterpaper]{article}
\usepackage{aaai17}
\usepackage{times}
\usepackage{helvet}
\usepackage{courier}
\usepackage[english]{babel}
\usepackage{blindtext}

\usepackage{amssymb}
\usepackage{graphicx}

\usepackage{url}

\usepackage{textcomp}
\usepackage[capitalise]{cleveref}
\usepackage{nth}
\usepackage[usenames]{color}

\usepackage{tikz}
\usetikzlibrary{fit, positioning, shapes, arrows, intersections}

\graphicspath{{./figures/}}

\usepackage{array}
\newcolumntype{C}[1]{>{\centering\let\newline\\\arraybackslash\hspace{0pt}}m{#1}}

\usepackage{amsmath}
\usepackage{subfigure}
\usepackage{multirow}

\begin{document}
\title{A Video-Based Method for Objectively Rating Ataxia}
\author{Anonymous authors
}
\maketitle

\begin{abstract}

For many movement disorders, such as Parkinson's and ataxia, disease
progression is usually assessed visually 
by a clinician according to a numerical rating scale,
or using questionnaires.
These tests are subjective, time-consuming, and must be administered
by a professional. We present an automated method for
quantifying the severity of motion impairment in patients with ataxia, using only
video recordings. We focus on videos of the finger-to-nose test, a common
movement task used to assess ataxia progression during the course of routine
clinical checkups.

Our method uses pose estimation and optical flow techniques to track the
motion of the patient's hand in a video recording. We extract features that
describe qualities of the motion such as speed and variation in performance.
Using labels provided by an expert clinician, we build a supervised
learning model that predicts severity
according to the Brief Ataxia Rating Scale (BARS). Our model achieves a mean absolute
error of $0.363$ on a $0-4$ scale and a prediction-label correlation of
$0.835$ in a leave-one-patient-out experiment. The accuracy of our system is comparable to the
reported inter-rater correlation among clinicians assessing the finger-to-nose exam using
a similar ataxia rating scale. This work demonstrates the feasibility of using
videos to produce more objective and clinically useful measures
of motor impairment.

\end{abstract}

\section{Introduction}

Over $9$ million people globally are affected by movement disorders, and this
number is projected to double in the next few decades
\cite{bach2011projected}.
Quantifying the severity of motor incapacity is useful
in monitoring the progression of these diseases and measuring the
effectiveness of treatments. Estimates of motor incapacity are most commonly 
made via questionnaries, or visual assessments combined with numerical rating scales
\cite{goetz2008movement,schmahmann2009development,schmitz2006scale}. While
many of these rating scales have been shown to be useful
\cite{schmitz2006reliability,schmahmann2009development,schmitz2006scale}, the
tests are subjective, and must be administered by an experienced clinician. An automated,
objective measurement of severity could open the door to more
consistent and fine-grained evaluations, and even self-administered testing.

We present an automated video-based method for quantifying the severity of
motor impairment in patients with ataxia. The term \textit{ataxia} describes a heterogenous group of neurodegenerative
diseases characterized by gait incoordination, involuntary
extremity and eye movements, and difficulty in articulating speech \cite{schmahmann2004disorders,klockgether2010sporadic}. The severity of ataxia is
often assessed using motor function tests such as the finger-to-nose maneuver,
in which a patient alternates between touching his/her nose and the
clinician's outstretched finger. A neurologist observing the patient's action rates the
disease severity on a numeric scale such as the Brief Ataxia Rating Scale
(BARS) \cite{schmahmann2009development}. These rating scales often consider aspects of the
patient's movement such as speed, smoothness, and accuracy. This process typically happens during scheduled clinical visits, which can
be time-consuming for the patient as well as the neurologist. Furthermore, this process ultimately
produces a subjective and coarse-grained rating of the patient's degree of
motor impairment.

Our work focuses on automatically producing a rating of motor impairment
directly from a video recording of a patient performing the finger-to-nose test. Our
dataset consists of videos taken with a handheld camera during clinical
visits. Some of these videos
contain camera movements such as panning and zooming. The dataset also contains
considerable variation in viewing angle and lighting conditions. While these
characteristics present many challenges, the video quality is representative of what one might expect from
clinical or home settings. Therefore, it is important for our system to be robust to
 the issues raised by quality. Another challenge is that patients with
the same severity rating could be impaired in different aspects of their motion.

Our method combines optical flow and pose estimation techniques to track the
location of the patient's wrist and head in each video. We extract features
from this motion signal, and use them in a learning algorithm to build a model
that predicts the severity rating of the patient's action on the BARS scale.
We achieve a mean absolute error of $0.363$ on a $0-4$ BARS scale and a
Pearson's correlation of $0.835$ relative to the score provided by an
experienced neurologist specializing in ataxia. The accuracy of our model compares
favorably to the inter-rater consistency of humans that has been reported for the finger-to-nose maneuver in a similar
ataxia rating scale \cite{burk2009comparison,weyer2007reliability}. This work demonstrates the feasibility of an automated
method for assessing the severity of motor impairment in ataxic patients. Such a
system could be useful in clinical or even home settings by allowing more
frequent and more objective assessment of the disease. An objective assessment
could also be useful for reducing observer bias in clinical trials.

\section{Related Work}
\subsection{Ataxia Rating Scales}
Ataxias are a group of degenerative
movement disorders that are characterized by incoordination of the extremities,
eyes and gait \cite{schmahmann2004disorders,klockgether2010sporadic}. Since their
introduction, quantitative rating scales have been the gold standard for ataxia
severity assessment. Some examples of the scales include the International
Cooperative Ataxia Rating scale (ICARS), the Scale for the Assessment and
Rating of Ataxia (SARA) and the Brief Ataxia Rating Scale (BARS)
\cite{trouillas1997international,schmitz2006scale,schmahmann2009development}.
These scales require an expert clinician to visually assess the qualities of the
patient's movements and determine a numerical rating, and can be time-consuming. In 2009, the BARS was
developed as a scale that is sufficiently fast and accurate for clinical
purposes. Nonetheless, as the designer of the BARS says, an objective,
fine-grained method for assessing ataxia is still ``sorely
needed'' \cite{schmahmann2009development}. We believe that this work is the
first step towards such a method.

\subsection{Human Motion Analysis}

The use of camera systems to measure and detect pathological movement is
well-studied for many diseases \cite{sutherland2002evolution,galna2014accuracy}. These systems
typically utilize multiple camera setups along with passive or active markers
and other sensing methods such as electromyography to track the motion of the
subject \cite{muro2014gait,kugler2013automatic,lizama2016use}. These systems
are expensive and complicated, must be operated by expert technicians,
and often require experts to interpret the data.
Other simpler systems utilize specialized hardware such as depth
cameras \cite{galna2014accuracy} or accelerometers \cite{mancini2012isway} to
measure specific pathological movements. While these kinds of hardware are
becoming more readily available, monocular consumer camera-based systems are
still easier to set up and operate in clinical settings. In this work, we
focus on monocular consumer-quality videos, since their ubiquity could allow
for widespread applications of our technique in both clinical and home
settings.

To the best of our knowledge, this is the first work that uses monocular video
recordings to automatically assess the severity of a neurological movement
disorder. Most existing approaches rely on more specialized hardware
\cite{mancini2012isway,sutherland2002evolution}, and solve different problems:
they measure movements but do not produce a rating of disease severity
\cite{galna2014accuracy}, or they focus on differentiating heathy and impaired
patients \cite{fazio2013gait,weiss2011toward}. One exception is the work by
\citeauthor{giuffrida2009clinically}, which produces a Unified Parkinson's
Disease Rating Scale rating from inertial sensor data
\cite{giuffrida2009clinically,goetz2008movement}. Our work focuses on a
different disease and different input data, so a direct comparison would not
be meaningful.

\subsubsection{Human Motion Analysis in Monocular Videos}
Human analysis in monocular videos often uses pose estimation as the initial step.
Human pose estimation aims to localize a human's joints in images or
videos. State-of-the-art pose estimation systems are effective at localizing joints even when they are occluded
by objects or other parts of the body. We use the publicly available implementation of
a convolutional neural network-based pose estimator \cite{wei2016convolutional} to track the hand of the patient in our videos.

More generally, motion analysis in monocular videos often relies on optical flow. Optical
flow measures the relative motion between the pixels in one frame and another.
Optical flow algorithms can be divided into sparse and dense categories
\cite{bruhn2005lucas}. Sparse flow methods
\cite{lucas1981iterative,shi1994good}, track only a sparse set of salient
points in the video while dense flow methods \cite{horn1981determining}
measure motion at every pixel in the frame. We use sparse optical flow to
track background points for video stabilization, and use dense optical flow
\cite{brox2009large} to refine the output of our pose estimator and to improve
the tracking of the patient's hand in each video.

\section{The BARS Dataset}\label{sec:dataset}
\begin{figure}
    \centering
    \includegraphics[scale=0.35]{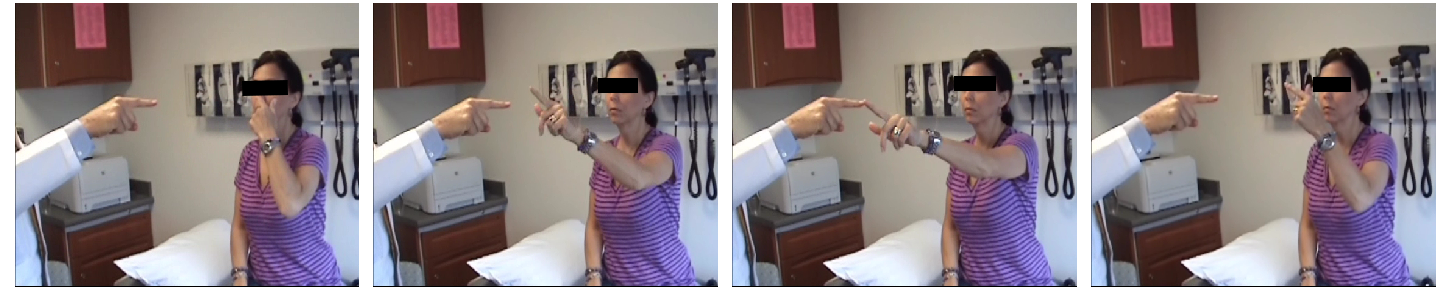}
    \caption{A finger-to-nose exam. The patient alternates between touching his/her nose and the clinician's outstretched finger. Each video in our dataset
    contains at least two repetitions of this action.} 
    \label{fig:vidPreview}
\end{figure}

In order to train a system to predict the BARS rating from a video, we use a
dataset of videos of ataxic patients. Our dataset consists of $92$ videos. The videos show $44$ distinct subjects
performing several repetitions of the finger-to-nose test with one hand at a
time (\Cref{fig:vidPreview}). The videos were shot during clinic visits with handheld cameras.
All videos were labeled by an experienced neurologist specializing in ataxia,
according to a version of the BARS that uses half-point
increments. For several patients, there are two videos for
each hand. The scores were
assigned separately for each hand, so a patient might have different ratings
for their right and left hands.

\begin{figure}
    \centering
    \includegraphics[scale=0.5]{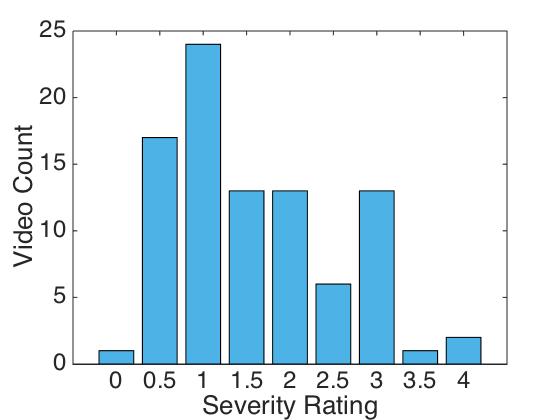}
    \caption{The distribution of severity ratings in the BARS dataset.} 
    \label{fig:fullHist}
\end{figure}

Eight videos were shot with the patient's hand moving almost directly towards the
camera. This unusual viewpoint makes it difficult for modern computer vision techniques to discern motion along the
nose-finger axis. These videos were excluded from the dataset, leaving us with $84$ videos of $40$ patients. \Cref{fig:fullHist} shows the
distribution of the severity ratings for the remaining $84$ videos.

Some videos contain occasional panning and zooming. We adjust for this by
using sparse optical flow to track points in the background, and then
estimating a similarity transformation from the tracked points to stabilize
each frame \cite{shi1994good,lucas1981iterative}. Because of the low number of
salient background points in some of the videos, only 61 videos were successfully
stabilized. We processed the remaining videos without stabilization.


\section{Feature Extraction}\label{sec:features}
From the videos in the BARS dataset, we extract features that
quantify the motion characteristics of each patient's active hand. We use pose
estimation and optical flow to track the location of the patient's wrist and
head, segment the motion signal into cycles, and compute motion features from
the signal segments. This process is summarized in \Cref{fig:featextraction},
and explained in detail in the following sections.

\subsection{Head and Wrist Tracking}
\begin{figure}[h!]
\begin{tikzpicture}
\tikzset{
smallblock/.style= {draw, rectangle, align=center,text width=3.2cm, inner sep=1mm, minimum height=1cm, node distance=0.7cm},
block/.style= {draw, rectangle, align=center,text width=4.5cm, inner sep=1mm, minimum height=1cm, node distance=0.7cm},
largeblock/.style= {draw, rectangle, align=center,text width=5.2cm, inner sep=1mm, minimum height=1cm, node distance=1cm},
line/.style = {draw, -latex'}
}

\node(poseest)[smallblock]{Run pose estimator\\ on each frame};

\node (A) [above right=0.4cm and -3.4cm of  poseest]  {\textbf{Track patient's head and active wrist}};

\node(optflow)[block, right=0.5cm of poseest]{Compute dense optical flow\\ between all adjacent frames};
\node(tempreg)[largeblock, below right=0.6cm and -2cm of poseest]{Enforce temporal continuity\\ of pose estimates\\ using optical flow};
\node[fit=(A)(poseest)(optflow)(tempreg), dashed,draw,inner sep=0.2cm] (Box)   {};
\node(relloc)[largeblock, below=0.8cm of tempreg]{Compute wrist location relative to head location};
\node(seg)[largeblock, below=0.6cm of relloc]{Segment wrist location signal\\ into cycles};
\node(compfeat)[largeblock, below=0.6cm of seg]{Compute motion features};

\path[line](poseest)--(tempreg);
\path[line](optflow)--(tempreg);
\path[line](tempreg)--(relloc);
\path[line](relloc)--(seg);
\path[line](seg)--(compfeat);

\end{tikzpicture}
\caption{Our feature extraction algorithm involves tracking the location of
the patient's active wrist, segmenting the location signal, and then computing
motion features.}
\label{fig:featextraction}
\end{figure}
\subsubsection{Head and Wrist Location Estimates Using Pose Estimation}
As a first step towards extracting features, we obtain the location of the
moving hand in each video frame. The quality of the videos, which are
representative of what might be captured in practice, make this a difficult
task. Most of the videos have low contrast, harsh lighting, and significant
motion blur, making local appearance-based tracking techniques such as sparse
point tracking \cite{lucas1981iterative} or mean-shift \cite{comaniciu2000real}
ineffective. In addition, there is significant variation in viewing angle
across videos, causing simple hand-recognition detectors to fail. We instead
rely on a more complex pose estimation system that calculates the likely
location of each body part in an image by leveraging information about the
configuration of the entire body. We use a state-of-the-art pose estimation
system based on convolutional neural networks \cite{wei2016convolutional} due to the availability of the code, even though the system was designed for images rather than videos.
The system is capable of predicting the location of several body joints, such as the wrist
and the top of the neck (bottom of the head). We use the relative location of the wrist to the top 
of the neck to approximate the location of the patient's hand relative to his/her head.

Using the pose estimator ``out of the box'' did not produce accurate wrist
location estimates. This is because the videos in our dataset contain
poses that are not well-represented in the MPII Human Pose dataset
\cite{andriluka20142d}, the dataset that the pose estimator was trained on. To
address this, we fine-tune a network on a new annotated dataset of individuals
performing the finger-to-nose-test, none of whom appear in the BARS dataset. Our dataset
consists of $626$ images from $26$ distinct individuals; $507$ images are taken from videos of $15$ healthy individuals, and $119$ images are downloaded from
internet resources on ataxia. We annotated the locations of joints in our dataset and fine-tuned the
network for $6,000$ iterations with a batch size of $8$.

\begin{figure}
	\centering
	\includegraphics[scale=0.25]{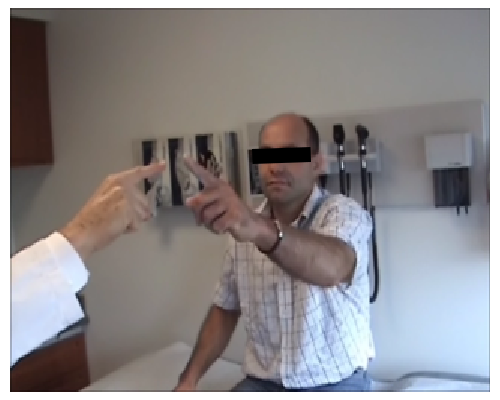}
	\includegraphics[scale=0.25]{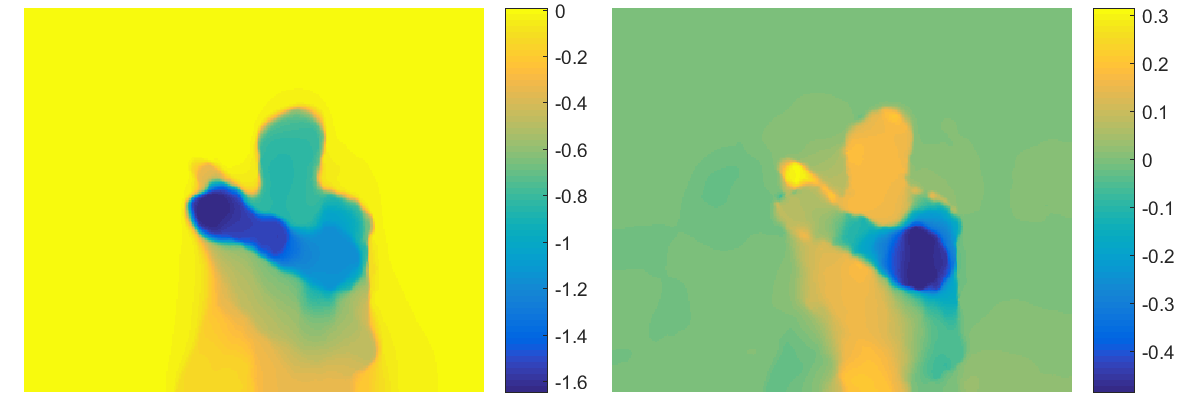}
	\caption{Flow field ($x$-direction, center; $y$-direction, right) computed for the video frame on the left.}\label{fig:flow}
\end{figure}

\subsubsection{Temporal Regularization Using Optical Flow}
\begin{figure}
	\centering
	\includegraphics[scale=0.22]{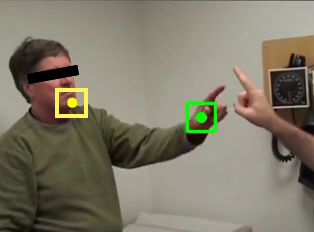}
	\includegraphics[scale=0.22]{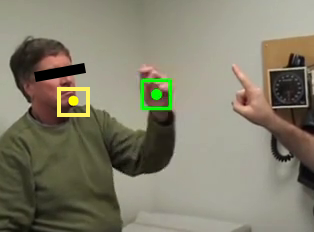}
	\includegraphics[scale=0.22]{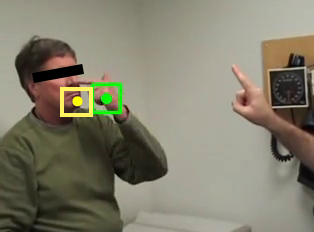}
        \caption{An example of our tracking results. The locations of the wrist and the bottom of the head are shown.}
        \label{fig:track}
\end{figure}

The pose estimator we use is designed for
images, and does not encode temporal continuity between neighboring video
frames. We enforce this continuity by 
temporally smoothing the joint estimates for each frame, using
the estimates from neighboring frames and dense optical flow as described in \cite{charles2014upper}. We
further improve the smoothed wrist location estimates by constraining the
estimates to fall within the fastest-moving region, which we assume to contain the patient's active hand. We determine these regions
by computing dense trajectories from flow spanning multiple frames as described in
\cite{sundaram2010dense}, and then selecting the trajectories with the highest
amount of motion over the course of the video. An example of our tracking results is shown in \Cref{fig:track}.

\subsection{Cycle Segmentation}
\begin{figure}[h]
		\subfigure[][Wrist position] {
			\centering
			\includegraphics[scale=0.5]{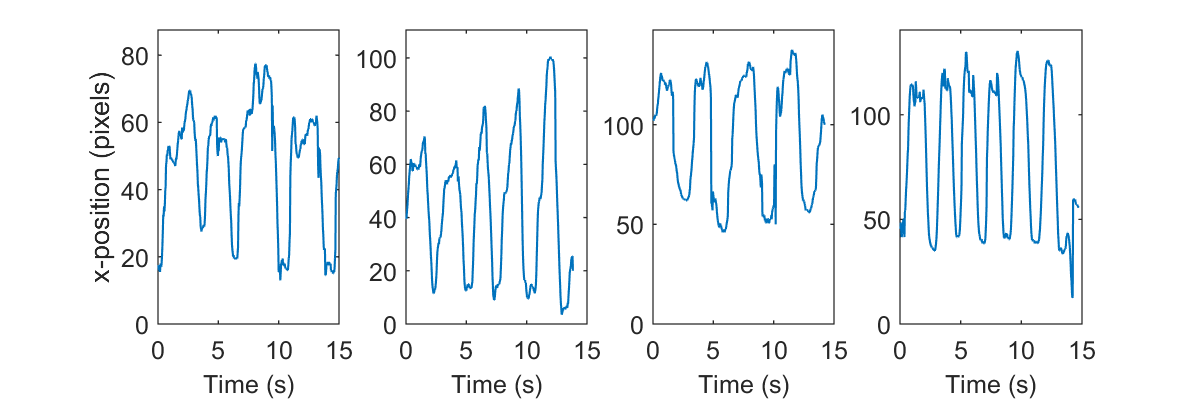}
                    }
~
		\subfigure[][Segmented wrist \newline position] {
			\centering
			\includegraphics[scale=0.5]{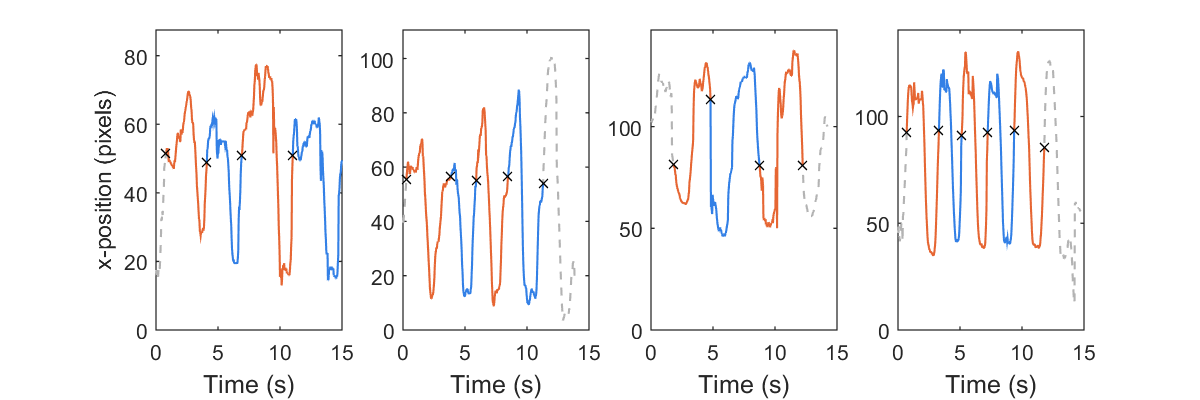}
                    }
		
	\caption{ (a) Wrist $x$-position signals (relative to the head) for four different patients that were
	each given a severity rating of $2$. The signals differ noticeably in shape and frequency.  (b) Segmented wrist $x$-position signals. The start of each cycle is marked with an $\times$. Portions of the signal that did not fall within a full cycle (in light gray) were discarded. }\label{fig:wristMotionSigs}
\end{figure}

\Cref{fig:wristMotionSigs}a shows examples of the wrist $x$-position signal (relative to head) produced by our tracking algorithm.
Each signal contains a varying number of repetitions of the finger-to-nose action.
To account for this, we assess the motion characteristics of each cycle independently.
We first segment the wrist location signal in time. The automated algorithm takes
as input the position of the patient's wrist relative to his/her head, and attempts to segment cycles beginning when the wrist is halfway between the patient's head and the doctor's finger. For robustness against tremors and noise around the midpoint, we use
hysteresis thresholding, a common thresholding technique used in signal processing, to detect the forward and backward portions of the
action. The threshold for the forward portion is $60\%$ of the way
from the patient's head to the doctor's finger; the backward threshold is
$40\%$. We approximate the location of the patient's head using the minimum of the wrist position signal, and the location of the doctor's finger using the maximum.
Since the beginning and the end
of the videos often include unrelated motions, we only compute the head and
doctor's finger positions from the middle half of the signal. We define a
cycle to be finger-nose-finger or nose-finger-nose based on which designation
produces the higher number of cycles in a video, and exclude any portions of
the signal that do not fall within a complete cycle. The results of applying
our cycle segmentation algorithm to the signals is shown in
\Cref{fig:wristMotionSigs}b.

\subsection{Motion Features}
From these wrist location signal segments, we extract 10 features. The features are
designed based on the BARS guidelines for rating the finger-to-nose test in
half-point increments. For example, the guidelines state that patients with a
severity rating of $0.5$ show slowed movement, and patients with a severity
rating of $1$ show oscillating movements. As the severity rating increases,
patients should show increasing variation of performance within the same
examination, as well as increasing amounts of dysmetria (a lack of
coordination).

\subsubsection{Average Cycle Duration}
As described in the BARS guidelines, healthy patients are able to complete
each cycle of the finger-to-nose test more rapidly than impaired patients. We
compute the average time it takes for each patient to complete a cycle, as
well as just the nose-to-finger and finger-to-nose portions of the cycles. We
use the logarithms of these three values in our feature vector, because we
hypothesize that a difference in cycle length at low severities is more
discriminative than the same difference at high severities.

\subsubsection{Number of Direction Changes}
We capture the amount of oscillation by counting the number of times the wrist
changes direction during the finger-to-nose action. We do this by counting the
number of sign changes in the first derivative of the wrist's $x-$ and $y-$position signals. Our features include the raw counts for both signals as well
as the counts normalized by the total number of cycles in the video. These
features also describe the patient's degree of dysmetria, since patients who
are more incoordinated have difficulty controlling the trajectory of their
hand during the test.

\subsubsection{Variation in Cycle Duration} 
Another important characteristic of ataxic movement is variation of
performance. Patients with more severe ataxia are unable to perform the
finger-to-nose action in a consistent manner. We capture this by computing the
standard deviations of the full cycle times, and the nose-to-finger and
finger-to-nose times.

\section{Model}
Using the features described in the previous section, we train a linear
regression model to predict the BARS severity rating for each video. Though the
BARS rating is not necessarily a linear function of our feature space, we prefer to use
a simpler model to avoid overfitting on our limited dataset. We use the LASSO technique,
a method for linear regression that includes a regularization term to help
with feature selection \cite{tibshirani1996regression}. We use
cross-validation to select the regularization parameter, and round the
predicted rating to the nearest valid BARS severity rating (which goes in half
point increments from 0 to 4).

\section{Results} \label{sec:learning-result}
We used leave-one-patient-out cross
validation to test our models. This approach allows us to train each model
with the maximum amount of data, and evaluate the model's performance on a
patient that it has not seen yet. Our models predicted the severity rating of each test video with a mean absolute error of $0.363$ on a $0-4$ BARS scale. The prediction-label
Pearson's correlation was $0.835$.

\subsection{Prediction Error}
\begin{figure}[h]
\centering

		\subfigure[][Distribution of \newline absolute errors] {
			\centering
			\includegraphics[scale=0.28]{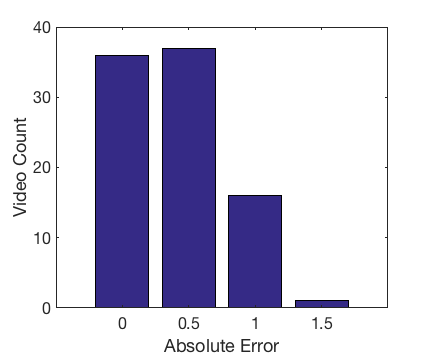}
                    }
~
		\subfigure[][Amount of error per ground truth rating] {
			\centering
			\includegraphics[scale=0.28]{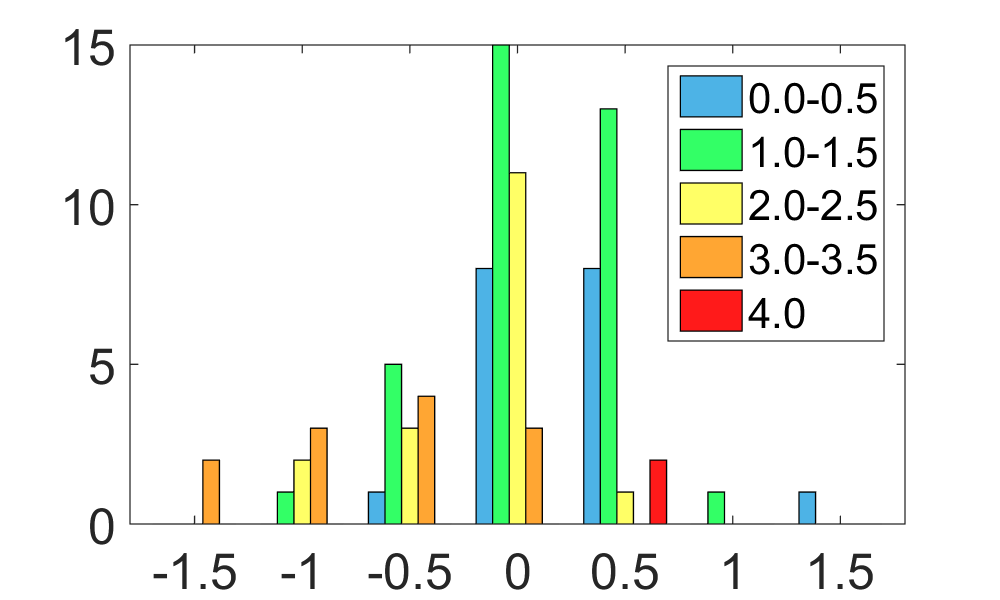}
                    }
		
	\caption{Distribution of errors in our leave-one-patient-out
	evaluation.}
	\label{fig:errordist}
\end{figure}

Most of our models' prediction errors are less than one level on
the BARS (88.1\%). \Cref{fig:errordist} shows the
distribution of errors, as well as the error for each ground truth severity
rating. Only three videos produced an absolute error greater than 1 point. In
one of those videos, the patient stopped in the
middle of the exam to ask the clinician a question. This pause was interpreted
by our system as a long average cycle duration, which caused our system to overestimate the rating by $1.5$. 
In the the other two
videos, the patient exhibited high amounts of
oscillation, but performed each finger-to-nose cycle fairly quickly, causing an underestimation by $1.5$. These
errors might be explained by the importance of the average cycle time feature
in our models, which received consistently high weights in the regressions. We
discuss this further in the Limitations section.

\subsection{Comparison with Human Performance}
\label{sec:humperf}
Our system is the first attempt at automatically producing a rating of ataxia
severity from video recordings. Thus, we compare the performance of our system to
that of human raters using an ataxia rating scale, the current gold standard
for assessing ataxia severity. Among human raters, the ground truth rating of 
ataxia severity is not well-defined, and thus the concept of error is less meaningful.
We instead compare the prediction-label correlation of our system to the inter-rater correlation of humans. 

To the best of our knowledge, no assessments of
human consistency in rating the finger-to-nose maneuver in the BARS have been
published. However, the evaluation of human consistency in rating the
finger-to-nose maneuver has been reported for the Scale for Assessment and
Rating of Ataxia (SARA) in two studies \cite{burk2009comparison,weyer2007reliability}. \citeauthor{burk2009comparison} focus
on 96 Freidreich's ataxia patients, and \citeauthor{weyer2007reliability}
study 64 patients of diverse types of ataxia. Like the BARS, the SARA rates the finger-to-nose maneuver from $0$ (normal) to $4$ (at
which the patient is unable to complete the maneuver), but differs in the criteria for the intermediate levels \cite{schmitz2006scale}. 

The BARS ratings in our dataset use a half-point scale, while the SARA uses a
full-point scale. In order to make a fair comparison, we investigate two methods for converting our dataset to use a full-point scale.
\begin{enumerate}

\item Discarding Half-Point Videos\\
We train and test on only the full-point videos in our dataset. This
restricted dataset contains 50 full-point videos of 32 distinct patients. We
evaluate our model using leave-one-patient-out cross-validation as described
earlier.

\item Random Rounding of Half-Point Videos\\
Half-point ratings may be interpreted as cases where the rater is uncertain
about the appropriate integer rating. We randomly round each half-point
rating up or down, and then train and evaluate our model on the entire rounded
dataset. We repeat the rounding process $100$ times and report the average
error and prediction-label correlation.
\end{enumerate}

Both dataset conversion methods are expected to have a negative effect
on our system's performance; discarding videos significantly decreases
the size of our dataset, and randomly rounding video ratings introduces
noise into our labels. 
\Cref{tab:fullpt} compares our system's performance on full-point videos to
the inter-rater correlation of human raters reported in the literature. The inter-rater correlation is presented using Pearson's correlation coefficient by \citeauthor{burk2009comparison}, and the intra-class correlation coefficient \cite{shrout1979intraclass}  by
\citeauthor{weyer2007reliability}
\cite{burk2009comparison,weyer2007reliability}. 
Despite the challenges posed by our dataset conversion methods, our system achieved considerably higher correlation than the Pearson's correlation coefficient of $0.55$ reported by
\citeauthor{burk2009comparison} Our system performed comparably to the
intra-class correlation of $0.793$ and $0.848$ reported for the right and left hand respectively
\cite{weyer2007reliability}.

Our system had mean
absolute errors of $0.417$ and $0.441$ respectively on the full-point datasets. While these errors are
higher compared to our models built using the half-point scale, the the error
relative to the rating step size is smaller.

\begin{table}[ht]
  \centering
	\begin{tabular} {| C{2cm} | C{1.8cm} | C{1.5cm}    | C{1.5cm}  |} 
		\hline
                
                \multirow{2}{*}{Metric} &\multirow{2}{*}{Human} & \multicolumn{2}{c|}{Our System} \\ 
               \cline{3-4}
                & & Discard & Randomly Round \\ \hline \hline
                Pearson's Correlation          & 0.55 & 0.757 & 0.741  \\ \hline
                Intra-class Correlation  & 0.793 (right)\newline 0.848 (left)& 0.753 & 0.743 \\ \hline

  \end{tabular}
  \caption{ Comparison of our system's performance to that of human raters, in
terms of inter-class correlation as well as intra-class correlation. All
            correlations are computed with left and right hands combined unless otherwise noted.}
\label{tab:fullpt}
\end{table}

\section{Limitations}
Our results show that we can predict the BARS ratings with a prediction-label correlation that is comparable
to the published inter-rater correlation among humans. However, our method did fail on a few videos,
suggesting areas of improvement. For example, it underestimated the rating for a severity-$3$ patient who moved quickly but with inaccuracies and oscillations. Using a model more complex than linear regression
can help capture these more complicated cases, but a larger dataset would be necessary to avoid overfitting.

Our dataset contains a limited number of examples for the higher severities.
For instance, the videos with severity ratings of $3.5$ and $4$ were all of
the same patient. While our system performed reasonably well on these videos
(0.5 absolute error for all videos of this patient), it is unclear if our
system will generalize well for other severely impaired patients.
Our model could be made more generalizable by obtaining
more videos of severely impaired patients, making the dataset more evenly distributed over the range of BARS severity ratings.

The BARS ratings for our dataset were provided by a single neurologist. 
Obtaining BARS ratings from multiple neurologists would allow us to average the ratings and compute a ground truth label
that is less subject to personal bias. Furthermore, such ratings would allow us to make a more
direct comparison of our method to human raters.

\section{Conclusion}
We described a video-based method for quantifying the severity of motion
impairment, and used the method to predict the severity rating of ataxic
patients performing the finger-to-nose test described in the Brief Ataxia
Rating Scale (BARS). Our method uses convolutional neural network-based pose
estimation and optical flow to track the location of a patient's wrist and
head. This tracking method 
is robust to the variations in camera angle and lighting conditions present in our dataset. We
segmented the resulting wrist position signal into cycles, and extracted
features such as average cycle time and amount of oscillation from these
signal segments based on the BARS guidelines. We used these features to build
a linear regression model to predict the severity rating of the patient's
action. Our models had a mean absolute error of $0.363$ on a $0-4$ BARS scale
and a correlation of $0.835$ relative to the ratings assigned by a trained
neurologist. This performance is comparable to the inter-rater correlation
reported for humans using a similar ataxia rating scale.

We have demonstrated the feasibility of an automated system for assessing the
severity of ataxic patients.
Such a system could be used to improve the management of the disease in clinical
or home settings, or to reduce observer bias in clinical trials. We believe that our system is an important step towards the automatic evaluation
of movement disorders.

\bibliography{references}
\bibliographystyle{aaai}

\end{document}